\documentclass{article}

\usepackage[preprint,nonatbib]{neurips_2020}

\usepackage[utf8]{inputenc} 
\usepackage[T1]{fontenc}    
\usepackage{hyperref}       
\usepackage{url}            
\usepackage{booktabs}       
\usepackage{amsfonts}       
\usepackage{nicefrac}       
\usepackage{multirow}
\usepackage{microtype}      
\usepackage{color}
\usepackage[pdftex]{graphicx}
\usepackage{amsmath,amssymb}
\usepackage{amsthm}

\usepackage{wrapfig}
\newlength{\bibitemsep}
\newlength{\bibparskip}
\let\oldthebibliography\thebibliography
\renewcommand\thebibliography[1]{%
  \oldthebibliography{#1}%
  \setlength{\parskip}{\bibitemsep}%
  \setlength{\itemsep}{\bibparskip}%
}
\usepackage{thmtools}
\usepackage{thm-restate}

\newtheorem{problem}{Problem}

\declaretheoremstyle[%
  spaceabove=4pt,%
  spacebelow=0pt,%
  bodyfont=\normalfont\itshape,
  postheadspace=1em,%
]{mystyle} 
\declaretheorem[name={Definition},style=mystyle,
]{defn}

\usepackage{algorithm}[compatible]
\usepackage{algpseudocode}
\usepackage[export]{adjustbox}
\usepackage{rotating}

\title{Ray-based classification framework for high-dimensional data}
\author{
Justyna P.~Zwolak\\
National Institute of \\Standards and Technology \\
Gaithersburg, MD 20899\\
\texttt{jpzwolak@nist.gov} \\
\And
Sandesh S.~Kalantre\\
Joint Quantum Institute \\
University of Maryland \\
College Park, MD 20742 \\
\texttt{skalantr@umd.edu} \\
\And
Thomas McJunkin \\
Department of Physics \\
University of Wisconsin \\
Madison, WI 53706 \\
\texttt{tmcjunkin@wisc.edu} \\
\And
Brian J.~Weber \\
Institute for Mathematical Sciences \\
ShanghaiTech University \\
Shanghai, 201210, China\\
\texttt{bjweber@shanghaitech.edu.cn} \\
\And
Jacob M.~Taylor \\
Joint Center for Quantum Information \\ and Computer Science, \\ 
National Institute of Standards and Technology \\
Gaithersburg, MD 20899, USA\\
\texttt{jacob.taylor@nist.gov} \\
}

\begin{document}
\maketitle
\begin{abstract}
While classification of arbitrary structures in high dimensions may require complete quantitative information, for simple geometrical structures, low-dimensional qualitative information about the boundaries defining the structures can suffice. Rather than using dense, multi-dimensional data, we propose a deep neural network (DNN) classification framework that utilizes a minimal collection of one-dimensional representations, called \emph{rays}, to construct the ``fingerprint'' of the structure(s) based on substantially reduced information. We empirically study this framework using a synthetic dataset of double and triple quantum dot devices and apply it to the classification problem of identifying the device state. We show that the performance of the ray-based classifier is already on par with traditional 2D images for low dimensional systems, while significantly cutting down the data acquisition cost. 
\end{abstract}

\section{Introduction}
\label{sec:intro}
Deep learning, with its remarkable progress in recent years~\cite{AlexNet,Goodfellow-et-al-2016}, is ripe for applications in physics~\cite{carleo-RevModPhys.91.045002}. A particular instance having general applicability to physical problems is the classification of arbitrary convex geometrical shapes embedded in an $N$-dimensional space~\cite{PhysRevResearch.2.023169}. Having a mathematical framework to understand this class of problems and a solution that scales efficiently with the dimension $N$ is essential. With increasing effective dimensionality of the system,
including parameters and data, determining the geometry with measurements across the full parameter space may become prohibitively expensive. However, as we show, qualitative information about the boundaries defining the structures of interest may suffice for classification. 

We propose a new framework for classifying simple high-dimensional geometrical structures: {\it ray-based classification}. Rather than working with the full $N$-dimensional data tensor, we train a fully connected DNN using one-dimensional representations in $\mathbb{R}^N$, called ``rays'', to recognize the relative position of features defining a given structure. We position the boundaries of this structure relative to a point of interest, effectively ``fingerprinting'' its neighborhood in the $\mathbb{R}^N$ space. The ray-based classifier is motivated primarily by experiments, particularly those in which sparse data collection is impractical. Our approach not only reduces the amount of data that needs to be collected, but also can be implemented {\it in situ} and in an online learning setting, where data is acquired sequentially. 

We test the proposed framework using a modified version of the ``Quantum dot data for machine learning'' dataset~\cite{ml-data} developed to study the application of convolutional neural networks (CNNs) to enhance calibration of semiconductor quantum dot devices for use as qubits~\cite{Loss98-QCD}. Tuning these devices requires a series of measurements of a single response variable as a function of voltages on electrostatic gates. As the number of gates increases~\cite{Zajac16-SGA,Mukhopadhyay18-2DD}, heuristic classification and tuning becomes increasingly difficult, as does the time it takes to fully explore the voltage space of all relevant gates. The specific geometry of the response in gate-voltage space corresponds to the number and position of populated quantum dots, which is valuable information in the process of tuning of these systems.

Previous work has shown both theoretically~\cite{Kalantre17-MLD} and experimentally~\cite{Zwolak20-AQD} that an image-based CNN classifier for 2D volumes, \textit{i.e.}, solid images, combined with conventional optimization routines, can assist experimental efforts in tuning quantum dot devices between zero-, single- and double-dot states. Here, we consider a double- and triple-dot system. We show that using ray-based classification, the quantity of data required (and thus the time required) for identifying the state of the quantum dot system can be drastically reduced compared to an imaged-based classifier.

\section{A Framework for Ray-Based Classification} 
\label{sec:framework}
\begin{figure}
  \begin{minipage}[c]{0.45\textwidth}
    \includegraphics[width=\textwidth]{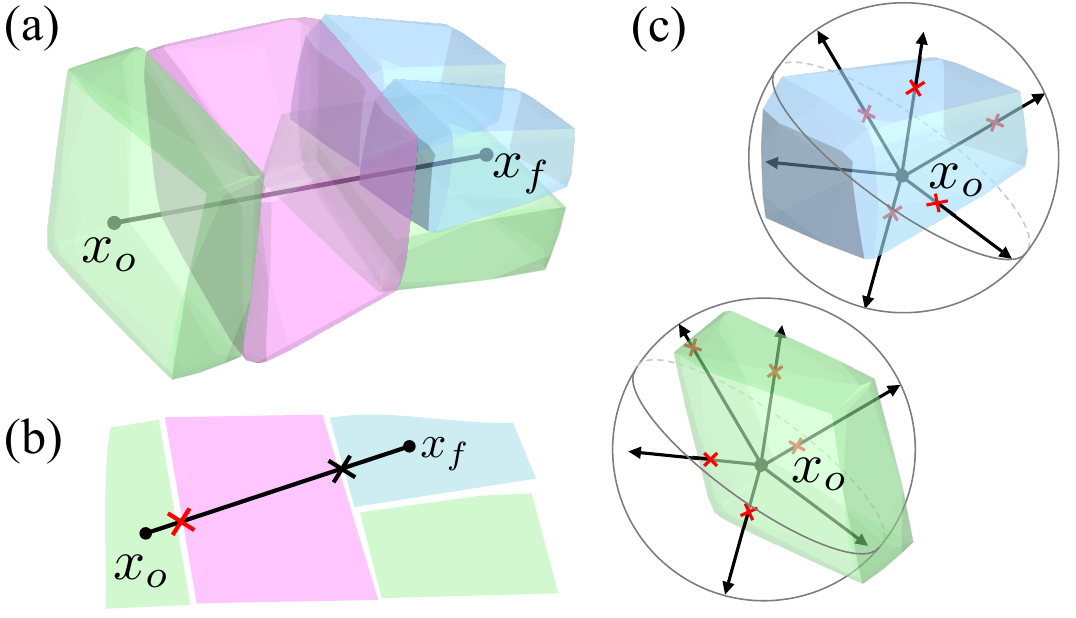}
  \end{minipage}\hfill
  \begin{minipage}[c]{0.5\textwidth}
    \caption{(a) Visualization of a ray $\mathfrak{R}(x_o,x_f)$ from $x_o$ to $x_f$ in $\mathbb{R}^3$. Different colors of polytopes represent different classes. (b) A side-view of the polytopes with two features marked along the ray. The red mark denotes a critical feature. (c) Visualization of the $M$-projection from point $x_o$ with $6$ rays (denoted by black arrows) for two different polytopes in $\mathbb{R}^3$. Note that both $M$-projections include a ray that does not have a critical feature.} \label{fig:rays-vis}
  \end{minipage}
  \vspace{-2pt}
\end{figure}

Consider Euclidean space $\mathbb{R}^N$ with its conventional 2-norm distance function $d$, and a polytope function $p : \mathbb{R}^N \rightarrow \{0,1\}$. The set of points where $p(x) = 1$ constitutes the boundary of a collection of polytopes. For example, a polytope function producing a square in $\mathbb{R}^2$ centered at the origin is $p(x_1,x_2)=\{1 \text{ if } |x_1|+|x_2| = 1; 0 \text{ elsewhere}\}$, where $(x_1, x_2)\in\mathbb{R}^2$. In our quantum dot applications a value of $p=1$ indicates the location where an electron is transferred in or out of a dot. 

\begin{defn}[Rays]\label{def:rays} 
	Given $x_o, x_f\in\mathbb{R}^N$, the {\bf\emph{ray}} $\mathfrak{R}_{x_o,x_f}$ emanating from $x_o$ and terminating at $x_f$ is the set $\{x\,|\,x = (1-t) x_o + t x_f, t \in [0,1]\}$ \emph{(see Fig.~\ref{fig:rays-vis}(a) for a depiction of a ray in $\mathbb{R}^3$)}.
\end{defn}
\vspace{-2pt}
In practical applications, rays have a natural granularity that depends on the system as well as the data collection density. For quantum dots, the device parameters define an intrinsic separation between critical features that gives the scale of the problem. We refer to granularity of rays in terms of pixels. 

To assess the geometry of a polytope enclosing any given point $x_o$, we consider a collection of rays of a fixed length $r$ centered at $x_o$. The rays are uniquely determined by a set of $M$ points on the sphere $\mathcal{S}^{N-1}_r$ of radius $r$ centered at $x_o$, $\mathcal{P}:=\{x_m\in\mathcal{S}^{N-1}_{x_o}(r)\,|\,1\le m \le M\}$. We call a set of $M$ rays, $\mathcal{R}_M:=\{\mathfrak{R}_{x_o,x_m}\,|\, x_m \in \mathcal{P}\}$, an {\bf\emph{$M$-projection}} (see Fig.~\ref{fig:rays-vis}(c) for visualization in $\mathbb{R}^3$).

\begin{defn}[Feature]
	Given a ray $\mathfrak{R}_{x_o,x_f}$ and a polytope function $p$, a point $x\in\mathfrak{R}_{x_o,x_f}$ is a {\bf\emph{feature}} if $p(x)=1$.
\end{defn}
\vspace{-2pt}
Figure~\ref{fig:rays-vis}(b) shows two features along a sample ray in $\mathbb{R}^3$. Features along a given ray define its {\bf\emph{feature set}}, $F_{x_o,x_f} := \{x\in\mathfrak{R}_{x_o,x_f}\,|\,p(x)=1\}$, with a natural order given by the 2-norm distance function $d:{x_o}\times F_{x_o,x_f} \rightarrow \mathbb{R}^+$. In general, $F_{x_o,x_f}$ could be empty. Using a decreasing {\bf\emph{weight function}} $\gamma:\mathbb{R}^+\rightarrow[0,1]$ we can assign a weight to each feature, effectively defining the {\bf\emph{weight set}} $\Gamma_{x_o,x_f}$ corresponding to its feature set $F_{x_o,x_f}$ as $\Gamma_{x_o,x_f} = \{\gamma(d(x,x_o))\, |\, x \in F_{x_o,x_f}\}$. The actual choice of function $\gamma$ needs be altered to fit the problem itself and can be considered another hyperparameter that can help optimize the machine learning process. For the quantum dot case, we chose $\gamma(n)=\nicefrac{1}{n}$. 

The assumption that the weight function $\gamma$ is monotonic in distance lets us define a ray's {\bf\emph{critical feature}} as the point $x \in F_{x_o,x_f}$ with highest (i.e., {\bf\emph{critical}}) weight $W_{x_o,x_f}=\gamma(d(x,x_o))$. If $F_{x_o,x_f}=\varnothing$, we put $W_{x_o,x_f}=0$. This allows us to ``fingerprint'' the space surrounding point $x_o$. 

\begin{defn}[Point fingerprint] 
	Let $x_o\in\mathbb{R}^N$ be a point from which a collection of rays $\mathcal{R}_M=\{\mathfrak{R}_{x_o,x_f^1},\dots,\mathfrak{R}_{x_o,x_f^M}\}$ emanate. The {\bf\emph{point fingerprint}} of $x_o$ is the $M$-dimensional vector consisting of the rays' critical weights: $\mathcal{F}_{x_o}=\big(W_{x_o,x_f^1},\,\dots,\,W_{x_o,x_f^M}\big)$.
\end{defn}

\begin{wrapfigure}[23]{R}{0.6\textwidth}
\vspace{-23pt}
\begin{minipage}{0.6\textwidth}
\begin{algorithm}[H]
\caption{Ray-based fingerprinting algorithm}
\label{alg:data_classification}
{\it {\bf Step 1.} Find $M$-projection centered at $x_o$ given $r$.}
\begin{algorithmic}[1]
\State {\bf Input:} $x_o$, $r$, a set $\mathcal{P}$ of $M$ points on the $(N-1)$-sphere
\State $m \gets 1$; $\mathcal{R}_M \gets$ empty list 
\For{$m=1$ to $M$}
    \State Find $m$-th ray $\mathfrak{R}_{x_o,x_f^m}$ and append it to the list $\mathcal{R}_M$.
\EndFor 
\State {\bf Return:} List of $M$ rays $\mathcal{R}_M$.
\end{algorithmic}
{\it {\bf Step 2.} Fingerprint $x_o\in\mathbb{R}^N$ using rays in $\mathcal{R}_M$ from Step 1.}
\begin{algorithmic}[1]
\State {\bf Input:} $\mathcal{R}_M$, $\gamma : \mathbb{R}^+ \rightarrow [ 0,1 ]$
\State $m \gets 1$; $\mathcal{F}_{x_o} \gets$ empty list 
\For{$m=1$ to $M$}
\State Find the feature set $F_{x_o,x_f^m}$.
\If {$F_{x_o,x_f^m} \neq\varnothing$}
    \State \begin{varwidth}[t]{0.87\linewidth} Identify the critical feature $x_i^m$, find $W_{x_o,x_f^m}$ and append  it to the list $\mathcal{F}_{x_o}$.\end{varwidth}
\Else 
    \State Append $0$ to the list $\mathcal{F}_{x_o}$.
\EndIf   
\EndFor
\State {\bf Return:} The point fingerprint vector $\mathcal{F}_{x_o}$.
\end{algorithmic}
\end{algorithm}
\end{minipage}
\end{wrapfigure}

This point fingerprint $\mathcal{F}_{x_o}$ of $x_o$ is the primary object of the ray-based classification framework. If sufficiently many rays in appropriate directions are chosen from $x_o$, the fingerprint is sufficient, at least in principle, to qualitatively determine the geometry of the convex polytope enclosing $x_o$. Due to the cost of experimental data acquisition, determining {\it how few} rays are sufficient for a machine learning algorithm to make this determination is of crucial importance. Looking to establish a correspondence between the fingerprint $\mathcal{F}_{x_o}$ of point $x_o$ and the class of the polytope enclosing this point, we define the following problem:

\begin{problem}
	Given a set of bounded and unbounded convex polytopes filling an $N$-dimensional space and belonging to $C$ distinct classes, $C \in \mathbb{N}$, and a point $x_o\in\mathbb{R}^N$, determine to which of the classes the polytope enclosing $x_o$ belongs.
\end{problem}

A solution to this problem in the supervised learning setting can be obtained by training a DNN with the input being the point fingerprint and the output identifying an appropriate class. The procedural steps for the proposed classification algorithm for $N$-dimensional data in the form of pseudocode are presented in Algorithm~\ref{alg:data_classification}.

\section{Experiments: Classifying Shapes in 2D and 3D}\label{sec:experiment}
\begin{figure}[b]
  \centering
  \includegraphics[width=0.94\linewidth]{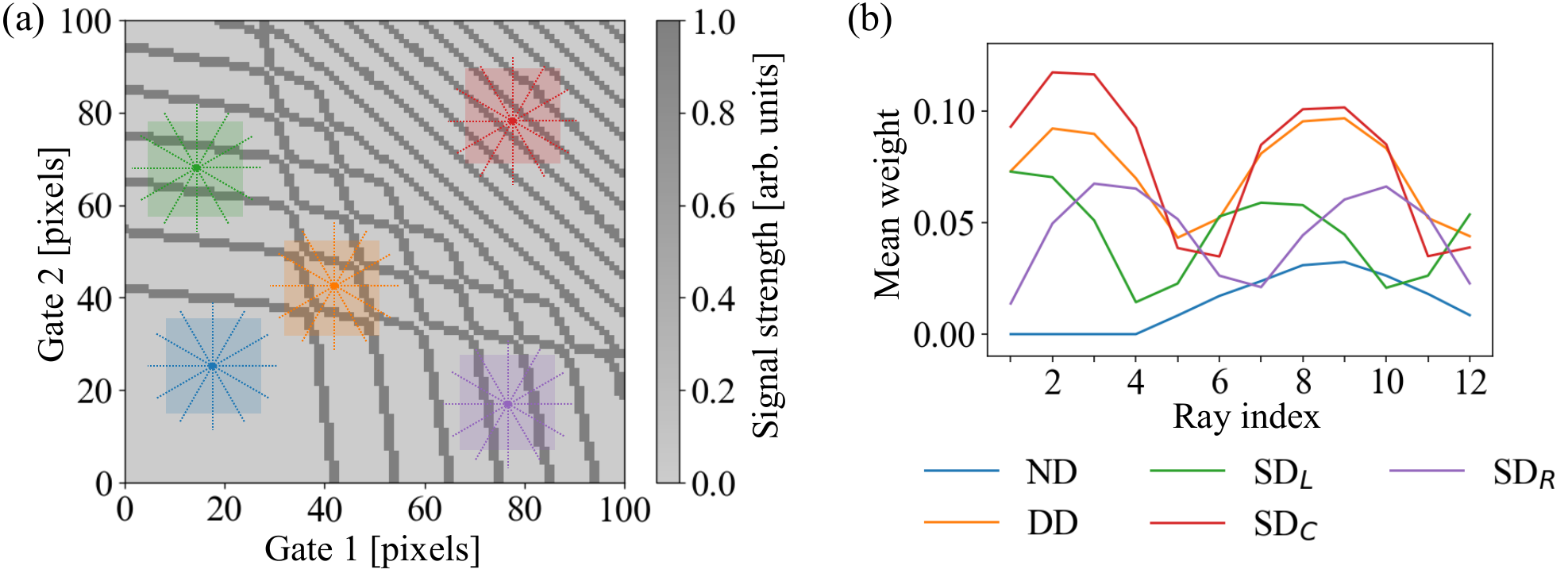}
  \vspace{-5pt}
  \caption{(a) A sample $2D$ map generated with the quantum dot simulator~\cite{Zwolak18-QLD} showing the different bounded and unbounded polytopes in $\mathbb{R}^2$ with $12$ evenly distributed rays overlaid on 2D scans like the ones used in Ref.~\cite{Zwolak18-QLD}. (b) The average trends of the fingerprints with $M=12$ rays. Fingerprints for SD$_L$ and SD$_R$ are out of phase, as expected from the the curvature of lines defining these states and SD$_C$ is shifted by $\nicefrac{1}{4}$ of the period). The colors (labels) are consistent between both panels.}
  \label{fig:sample_2D_map}
\end{figure}

The ray-based data is generated using a physics-based simulator of quantum dot devices~\cite{Zwolak18-QLD}. An example of a simulated measurement, like the ones typically seen in the laboratory, is shown in Fig.~\ref{fig:sample_2D_map}(a). The $x$ and $y$ axes represent a subset of parameters that can be changed in the experiments (here, gate voltages) and the curves where the signal strength is equal to 1 represent the device response to a change in electron occupation. The slopes of those lines correspond to the location of the quantum dots with respect to the gates. The device states manifest themselves by different bounded and unbounded shapes defined by these curves, as shown in Fig.~\ref{fig:sample_2D_map}(a). Previous work has confirmed the reliability of a dataset generated with this simulator for the case of a CNN used with $2D$ images, finding an accuracy of $95.9\,\%$ (standard deviation $\sigma=0.6\,\%$) over 200 training and validation runs performed on distinct datasets~\cite{Zwolak18-QLD}. Here, we use a modified version of this dataset, splitting the single-dot (SD) class into 3 distinct classes based on the dot location (Left, Center, Right) as suggested by experimentalists. No-dot (ND) and double-dot (DD) classes are unchanged.

To test the ray-based classification framework in 2D, we use $20$ realizations of $2D$ maps qualitatively comparable to the one shown in Fig.~\ref{fig:sample_2D_map}(a). Using a synthetic dataset allows us to systematically vary the length of the rays and their number. A regular grid of 1,369 points is used for sampling, resulting in a dataset of 27,380 fingerprints. We consider five datasets of $M$-projections, with $M=3,4,5,6,\mathrm{and }12$ evenly spaced rays. The ray length is varied between $10$ and $80$ pixels (where 30 pixels is the average separation between transition lines in the simulated devices). We ran 50 training and validation tests per combination of rays' number and length (with data divided 80:20). For testing, we generated a separate dataset based on three distinct devices. This allows us to both better determine the classification error for the most efficient number and length combinations of rays and to study the failure cases over the device layout (see Fig.~\ref{fig:perf_fc_vis}). 

\begin{figure}[t]
  \centering
  \includegraphics[width=0.95\linewidth]{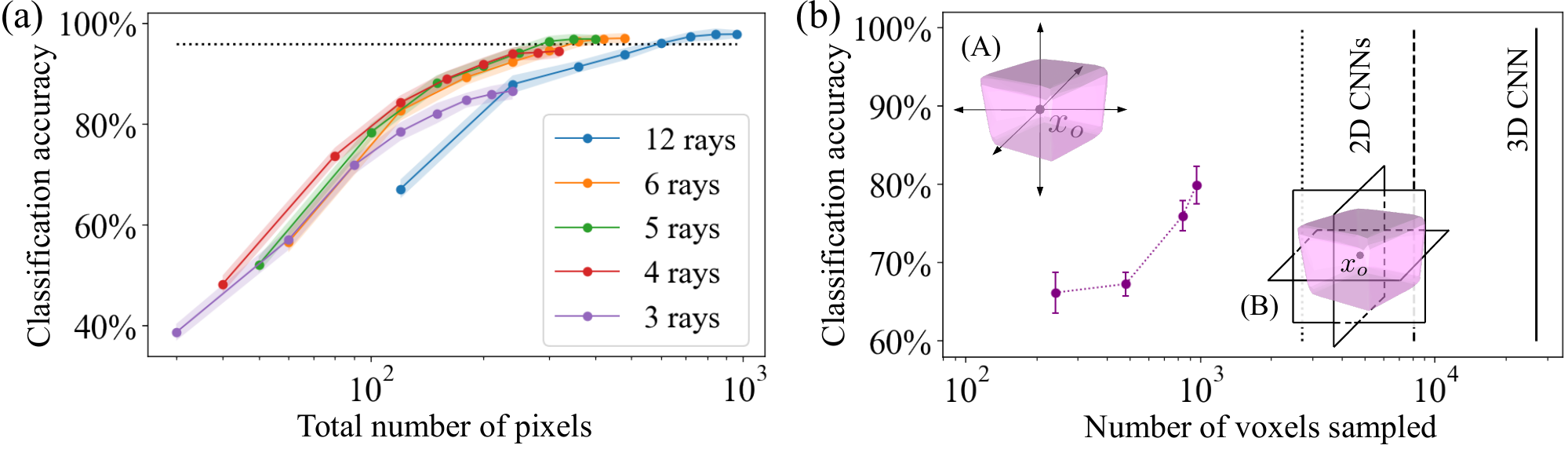}
  \caption{Classifier performance for varying numbers of rays as a function of the total number of (a) pixels measured and averaged over $N=50$ training runs for the double-dot system and (b) voxel number averaged over $N=10$ training runs for the triple-dot system. The black dashed line in (a) represents the benchmark from Ref.~\cite{Zwolak18-QLD}. The black vertical lines in (b) represent the minimum data requirements for CNN classifier with 3 orthogonal 2D slices (as depicted in insert (B), dotted line), large 2D scan (dashed line), and a full 3D CNN (solid line). Insert (A) shows the $M$-projection with 6 rays. In both panels, the connecting lines are a guide to the eye only and the $3\,\sigma$ confidence bands.}
  \label{fig:perf_var_rays}
\end{figure}

Figure~\ref{fig:perf_var_rays}(a) shows the performance of the ray-based classifier. The accuracy of the classifier increases with the total number of points measured for a fixed number or rays, as expected. However, for a fixed number of points, increasing the number of rays does not necessarily lead to increased accuracy. This is because with a fixed number of points and point density, increasing the number of rays naturally results in shorter rays. Rays shorter than the radius of the interior diameter of the shapes leads to empty feature sets, resulting in uninformative fingerprints. Increasing the number or size of hidden layers in the DNN does not further improve the accuracy (see Table A.1).

To test the proposed framework with triple-dot systems~\cite{Schroer2007}, we generated a dataset by sampling 17,576 fingerprints from a single simulated device with three dot gates. We varied the number of rays between 6 and 18, while keeping the length of the rays fixed at 60 voxels. For each configuration, we performed $N=10$ training and validation runs (with data divided 80:20). As shown in Fig.~\ref{fig:perf_var_rays}, the classifier accuracy improved from 66.2 \% ($\sigma$ = 0.3 \%) for 6 rays to 79.9 \% ($\sigma$ = 0.3 \%) for 18 rays.

\section{Summary}\label{sec:concl}
While our analysis is performed using simulated data, its true advantage becomes clear when put in the context of experiments. Depending on the resolution, measuring $12$ rays of length $80$ (total of $960$ data points) is equivalent to measuring the full $2D$ image ($900$ data points) as in Ref.~\cite{Zwolak20-AQD}. With 6 rays of length 60 pixels, only $360$ data points are needed, resulting in $60\,\%$ reduction of the data needed to obtain accuracy of $96.4\,\%$ ($\sigma=0.4\,\%$). The reduction for 3D data is even more significant.

In conclusion, we have defined a framework to generate a low-dimensional representation of geometrical shapes in a high-dimensional space. We have empirically shown that the ray-based framework is an effective solution for cutting down the measurement cost while preserving high-accuracy of classification on the quantum dot dataset. If the ray-based classifier were implemented in a scheme to tune the double dot, as in Ref. \cite{Zwolak20-AQD}, this reduction in data collection significantly improves its viability as a replacement to a hand-tuning scheme by a human operator. We expect this approach will find many related applications outside of the quantum dot domain.

\section*{Broader Impact}
The authors believe that research presented in this paper does not have ethical aspects. This work furthers the efforts to automate and scale up quantum dot-based quantum computing to larger and more impactful devices. Our approach can also be extended to setups involving the estimation of quantum states in solid-state and atomic experiments, as well as tuning and scalability of other quantum computing architectures. A ray-based approach may find use classifying point clouds if sufficiently ordered. Extending our technique to include intersection points beyond the first critical feature may allow for identification of non-convex shapes in higher dimensions.

\begin{ack}
This research was sponsored in part by the Army Research Office (ARO), through Grant No. W911NF-17-1-0274. S.K. gratefully acknowledges support from the Joint Quantum Institute (JQI) -- Joint Center for Quantum Information and Computer Science (QuICS) Lanczos graduate fellowship. The views and conclusions contained in this paper are those of the authors and should not be interpreted as representing the official policies, either expressed or implied, of the ARO, or the U.S. Government. The U.S. Government is authorized to reproduce and distribute reprints for Government purposes notwithstanding any copyright noted herein. Any mention of commercial products is for information only; it does not imply recommendation or endorsement by the National Institute of Standards and Technology.
\end{ack}


\newpage
\normalsize

\appendix
\renewcommand\thefigure{A.1} 
\renewcommand\thetable{A.1} 
\section*{Appendix}

\subsection*{Overview of failure modes}\label{sec:fc}
\begin{figure}[h]
\centering
\includegraphics[width=0.90\textwidth]{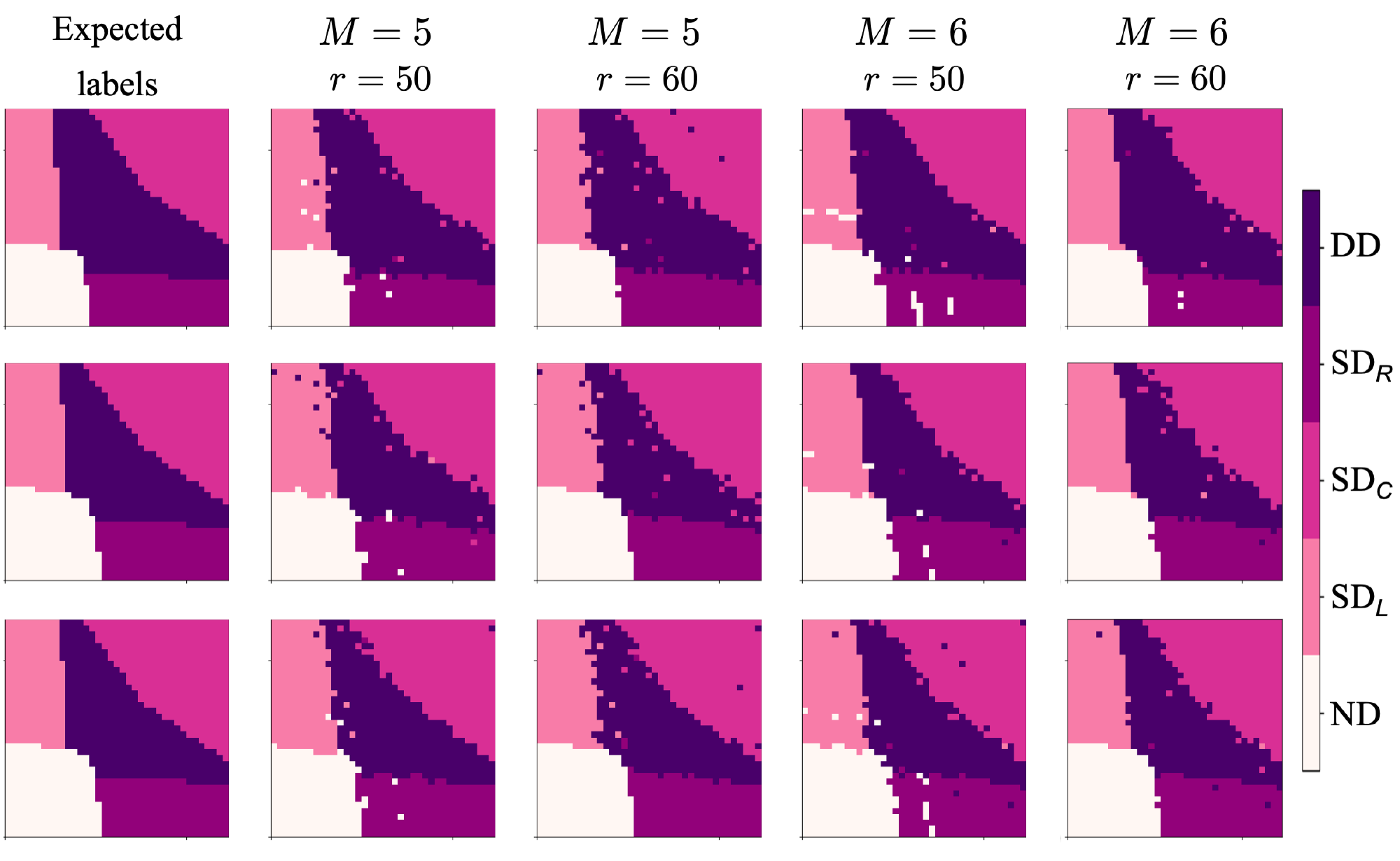}
\caption{Visualization of the failing cases overlaid on the test devices, with each row corresponding to a different device. The expected labels are shown in the leftmost column. The remaining columns are the ray-based classifier results for varying number of rays, $M$, and length of rays in pixels, $r$.}
\label{fig:perf_fc_vis}
\end{figure}

To better understand the failure cases of the ray-based classifier for the best rays' number and length combinations, we use three test datasets comprised of a regular grid of 1,369 points sampled over three devices distinct from those used for training and validation. Figure~\ref{fig:perf_fc_vis} visualizes the classification success on a stability diagrams like the one in Fig.~\ref{fig:sample_2D_map}(a). The test devices are shown as rows and the different rays' numbers and lengths combinations are shown as columns. As can be seen in columns two and four, with $r=50$ pixels the classifiers fails when a sampled point falls on the boundary lines for the polygons.

\subsection*{Analysis of DNN architectures}\label{sec:DNN}
In experiments presented in Sec.~\ref{sec:experiment} we use a DNN with 256, 128, and 32 neurons and the ReLU activation function for the fully connected hidden layers. The output layer consists of 5 neurons and the softmax activation function. We use the Adam optimizer ($\eta=10^{-3}$), and the sparse categorical cross-entropy loss function. We found that increasing the size or complexity of the DNN (in terms of more or bigger layers) does not improve the performance of the ray-based classifier. In fact, for certain rays' number and length combination (e.g., 6 rays of length 60 pixels), a smaller network would suffice to achieve the same accuracy (see Table~\ref{tab:DNN-variations}).

\begin{table}[h]
  \caption{Comparison of the varying DNN architectures for a fixed number and length of rays. For each network we report average accuracy, $\mu_{r}$ ($\%$), and standard deviation, $\sigma_{r}$ ($\%$), where $r$ denotes the length of the ray (in pixels), for $N=50$ iterations of training and testing.}
  \label{tab:DNN-variations}
  \centering
  \begin{tabular}{cllll}
    \toprule
    \multirow{2}{*}{DNN} & \multicolumn{2}{c}{5 rays} & \multicolumn{2}{c}{6 rays} \\
    \cmidrule(r){2-5}
     & $\mu_{50}\,\,(\sigma_{50})$ & $\mu_{60}$ ($\sigma_{60}$) & $\mu_{50}$ ($\sigma_{50}$) & $\mu_{60}$ ($\sigma_{60}$) \\
    \midrule
        64-32 & 93.6 (0.4) & 95.8 (0.4) & 94.5 (0.3) & 96.3 (0.3) \\
    128-64-32 & 94.2 (0.4) & 96.4 (0.4) & 94.6 (0.4) & 96.4 (0.4) \\
    256-64-32 & 94.2 (0.5) & 96.5 (0.4) & 94.7 (0.4) & 96.6 (0.3) \\
512-256-64-32 & 94.6 (0.4) & 96.5 (0.4) & 94.5 (0.4) & 96.3 (0.3) \\
    \bottomrule
  \end{tabular}
\end{table}


\begin{thebibliography}{10}
\bibitem{AlexNet}
A.~Krizhevsky, I.~Sutskever, and G.~E. Hinton.
\newblock Imagenet classification with deep convolutional neural networks.
\newblock In F.~Pereira, C.~J.~C. Burges, L.~Bottou, and K.~Q. Weinberger, editors, {\em Advances in Neural Information Processing Systems 25}, pages 1097--1105. Curran Associates, Inc. (2012).

\bibitem{Goodfellow-et-al-2016}
I. Goodfellow, Y. Bengio, and A. Courville.
\newblock {\em Deep Learning}.
\newblock \url{http://www.deeplearningbook.org}.
\newblock MIT Press (2016).

\bibitem{carleo-RevModPhys.91.045002}
G. Carleo, I. Cirac, K. Cranmer, L. Daudet, M. Schuld, N. Tishby, L. Vogt-Maranto, and L. Zdeborov\'a.
\newblock Machine learning and the physical sciences.
\newblock {\em Rev. Mod. Phys.} {\bf 91}, 045002 (2019).

\bibitem{PhysRevResearch.2.023169}
P. Rotondo, M.~C. Lagomarsino, and M. Gherardi.
\newblock Counting the learnable functions of geometrically structured data.
\newblock {\em Phys. Rev. Res.} {\bf 2}, 023169 (2020).

\bibitem{ml-data}
Quantum dot data for machine learning, 2018.
\newblock \url{https://doi.org/10.18434/T4/1423788}.

\bibitem{Loss98-QCD}
D.~Loss and D.~P. DiVincenzo.
\newblock Quantum computation with quantum dots.
\newblock {\em Phys. Rev. A} {\bf 57}, 120--126, (1998).

\bibitem{Zajac16-SGA}
D.~M. Zajac, T.~M. Hazard, X.~Mi, E.~Nielsen, and J.~R. Petta.
\newblock Scalable {Gate} {Architecture} for a {One}-{Dimensional} {Array} of {Semiconductor} {Spin} {Qubits}.
\newblock {\em Phys. Rev. Appl.} {\bf 6}, 054013 (2016).

\bibitem{Mukhopadhyay18-2DD}
U.~Mukhopadhyay, J.~P. Dehollain, Ch. Reichl, W.~Wegscheider, and L.~M.~K. Vandersypen.
\newblock A 2$\times$2 quantum dot array with controllable inter-dot tunnel couplings.
\newblock {\em Appl. Phys. Lett.} {\bf 112}, 183505 (2018).

\bibitem{Kalantre17-MLD}
S.~S. Kalantre, J.~P. Zwolak, S. Ragole, X. Wu, N.~M. Zimmerman, M.~D. Stewart, and J.~M. Taylor.
\newblock Machine learning techniques for state recognition and auto-tuning in quantum dots.
\newblock {\em npj Quantum Inf.} {\bf 5}, 1--10 (2019).

\bibitem{Zwolak20-AQD}
J.~P. Zwolak, T. McJunkin, S.~S. Kalantre, J.P. Dodson, E.R. MacQuarrie, D.E. Savage, M.G. Lagally, S.N. Coppersmith, M.~A. Eriksson, and J.~M. Taylor.
\newblock Autotuning of double-dot devices in situ with machine learning.
\newblock {\em Phys. Rev. Appl.} {\bf 13}, 034075 (2020).

\bibitem{Zwolak18-QLD}
J.~P. Zwolak, S.~S. Kalantre, X.~Wu, S.~Ragole, and J.~M. Taylor.
\newblock {QFlow} lite dataset: {A} machine-learning approach to the charge states in quantum dot experiments.
\newblock {\em PLoS ONE}, {\bf 13}, 1--17 (2018).

\bibitem{Schroer2007}
D. Schroer, A.~D. Greentree, L. Gaudreau, K. Eberl, L.~C.~L. Hollenberg, J.~P. Kotthaus, and S. Ludwig.
\newblock Electrostatically defined serial triple quantum dot charged with few electrons.
\newblock {\em Phys. Rev. B}{\bf 76} 075306 (2007)

\end{thebibliography}
\end{document}